\documentclass[letterpaper, 10 pt, conference]{ieeeconf}  % Comment this line out if you need a4paper

\IEEEoverridecommandlockouts                              % This command is only needed if 
\usepackage{graphics} % for pdf, bitmapped graphics files
\usepackage{amssymb}
\usepackage{graphicx}
\usepackage{caption}
\usepackage{amsmath}
\usepackage{hyperref}
\usepackage{url}
\usepackage{multirow}
\usepackage{booktabs}
\usepackage{cite}
\usepackage{booktabs,tabularx,array}
\usepackage{placeins}
\usepackage{bm}
\usepackage{comment}
\usepackage{xcolor}

\title{\LARGE \bf
Uncertainty-Aware Vision-based Risk Object Identification via 
Conformal Risk Tube Prediction
}

\author{%
Kai-Yu Fu \quad
Yi-Ting Chen$^{\dagger}$ \\
National Yang Ming Chiao Tung University \\
\thanks {$^{\dagger}$ Corresponding Author}
}

\begin{document}

\maketitle
\thispagestyle{empty}
\pagestyle{empty}

\begin{abstract}
We study object importance-based vision risk object identification (Vision-ROI), a key capability for hazard detection in intelligent driving systems.
Existing approaches make deterministic decisions and ignore uncertainty, which could lead to safety-critical failures.
Specifically, in ambiguous scenarios, fixed decision thresholds may cause premature or delayed risk detection and temporally unstable predictions, especially in complex scenes with multiple interacting risks.
Despite these challenges, current methods lack a principled framework to model risk uncertainty jointly across space and time.
We propose Conformal Risk Tube Prediction, a unified formulation that captures spatiotemporal risk uncertainty, provides coverage guarantees for true risks, and produces calibrated risk scores with uncertainty estimates.
%
% We further introduce a conformal prediction framework to provide coverage guarantees for the true risks and yield calibrated risk scores and uncertainty estimates.
%
% Specifically, we employ risk-category–aware calibrators that consider distinct characteristics to reduce confused calibration.
%
To conduct a systematic evaluation, we present a new dataset and metrics probing diverse scenario configurations with multi-risk coupling effects, which are not supported by existing datasets. 
We systematically analyze factors affecting uncertainty estimation, including scenario variations, per-risk category behavior, and perception error propagation.
Our method delivers substantial improvements over prior approaches, enhancing vision-ROI robustness and downstream performance, such as reducing nuisance braking alerts.
For more qualitative results, please visit our project webpage: \href{https://hcis-lab.github.io/CRTP/}{\textcolor{magenta}{https://hcis-lab.github.io/CRTP/}}
\end{abstract}
\section{INTRODUCTION}
%%% Vision-ROI background%%%
With over 1.19 million road traffic deaths annually~\cite{10665-375016}, improving the safety of intelligent driving systems (IDS) has been a longstanding goal in the community.
A key capability in this effort is visual risk object identification (Vision-ROI), which aims to localize potential hazards and estimate their associated risk levels or importance scores.
The community has explored a variety of approaches, including collision prediction~\cite{herzig2019spatiotemporalactiongraphnetworks,fang2023dadadriverattentionprediction,you2020CTA}, trajectory prediction and collision checking~\cite{Chandra_2019,malla2020titanfutureforecastusing,9577864}, object importance estimation~\cite{10.1007/978-3-540-88682-2_40,OHNBAR2017425,zeng2017agentcentricriskassessmentaccident,gao2019goalorientedobjectimportanceestimation, li2022importantobjectidentificationsemisupervised}, human gaze prediction~\cite{7789504, xia2018predictingdriverattentioncritical,pal2020lookingrightstuffguided,baee2021medirlpredictingvisualattention}, and behavior-based prediction~\cite{li2020learning3dawareegocentricspatialtemporal,li2020makedriversstopdrivercentric,Gupta_2024,xiao2023learningroadscenelevelrepresentations,10177972,pao2025potentialfieldsceneaffordance}.
In this paper, we study object importance-based Vision-ROI, where risk objects are defined by human annotators’ subjective assessment.
This formulation directly reflects human perception of driving risk and is a common supervision signal in driving datasets.

%%% Why we need uncertainty in Vision-ROI? %%%
Existing object-importance–based Vision-ROI approaches are largely deterministic, implicitly assuming reliable perception and stable scene dynamics.
However, real-world traffic environments are inherently uncertain due to factors such as occlusions, sensor noise, and incomplete observations that may conceal potential hazards.
% , such as occluded regions that may conceal potential hazards.
%
In these scenarios, fixed decision thresholds can lead to temporal boundary misalignment (i.e., premature or delayed risk detection and release) and fragmented predictions that flicker between risky and non-risky states.
Such behaviors are undesirable in safety-critical systems because they can produce unstable risk assessments near decision boundaries.
%
% These challenges highlight the need for IDS that explicitly model uncertainty rather than relying on deterministic predictions~\cite{}.
%
This gap motivates the need to develop uncertainty-aware Vision-ROI systems that adapt their risk assessment to the spatiotemporal complexity of the scene and operate reliably across diverse traffic configurations.~\cite{iso21448,unece2023validatingautomatedsrivingsystem,10879299}.

% Uncertainty-aware methods~\cite{10879299} have been shown to improve accuracy, interpretability, robustness to distribution shifts, and driving safety, while remaining efficient in tasks such as accident anticipation~\cite{Bao_2020}, trajectory prediction~\cite{huang2025cuqdsconformaluncertaintyquantification}, and planning~\cite{shao2024uncertaintyawarepredictionapplicationplanning}.
%
% However, existing object importance–based Vision-ROI approaches are largely deterministic and ignore predictive uncertainty; such overconfident outputs~\cite{gawlikowski2022surveyuncertaintydeepneural} may compromise safety~\cite{Khaitan2020SafePA,9815528}.
%
% 

\begin{figure}[t] 
  \centering
  \includegraphics[width=\columnwidth]{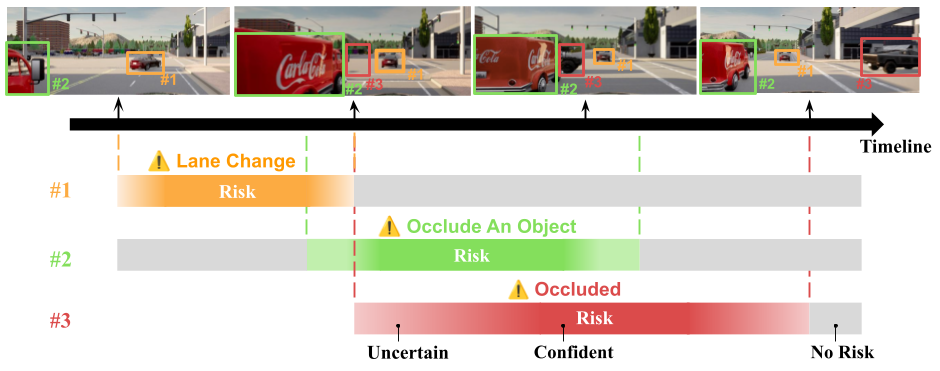}%
  \caption{\textbf{Risk Tube Prediction}. Our formulation models risk uncertainty jointly across space and time by representing potential hazards as spatiotemporal risk tubes. 
  In this example, the green-boxed truck (\#2) moves forward and may occlude part of the scene, creating the possibility of a hidden object (\#3) emerging from the occluded region.
  %
  % As the green-boxed truck (\#2) moves forward, it may create future occlusion, prompting anticipation of a hidden object (\#3) at the red-boxed location. 
  %
  Risk tubes illustrate how potential hazards evolve over time, while uncertainty is visualized through semi-transparent shading that gradually becomes opaque as observations reduce ambiguity and confidence increases.}
  \label{fig:RTP}
  % \vspace{-15pt}
\end{figure}

%%% What we need for uncertainty Vision-ROI? New Formulation (Risk Tube). %%%
To bridge this gap, we propose \textbf{Risk Tube Prediction} (fig.~\ref{fig:RTP}), an uncertainty-aware formulation for Vision-ROI that jointly models uncertainty over spatial extent and temporal horizon.
Instead of predicting risk for individual objects at a single time step, our formulation represents risk as a spatiotemporal tube that captures how potential hazards evolve over time.
% we study the challenges of incorporating uncertainty into object importance–based Vision-ROI and examine its impact on downstream tasks.
%
This representation is motivated by two key observations. First, risk in driving scenarios is inherently temporal: objects that are currently safe may become hazardous due to future interactions, motion patterns, or road topology. 
Predicting risk solely at the frame level therefore fails to capture the temporal development of hazards.
Second, uncertainty often arises not only from whether an object is risky, but also from where and when the risk may occur. 
Occlusions, partial observations, and complex multi-agent interactions can lead to ambiguity in both spatial and temporal localization of risks.
%
% In diverse scenarios with multiple interacting risks, hazards may shift in space and time, complicating reliable uncertainty estimation.
%
%
% Yet existing vision-based formulations still lack a way to indicate how uncertainty jointly evolves over space and time.
%
% To this end, we propose \textbf{Risk Tube Prediction} (fig.~\ref{fig:RTP}), an uncertainty-aware formulation for Vision-ROI that jointly models uncertainty over spatial extent and temporal horizon into a unified representation.
%
% By capturing spatiotemporal uncertainty and marginalizing variability in both location and timing, it yields more robust risk estimates.

%%% Existing uncertainty modeling are not sufficient. Why Conformal Prediction? %%%
To obtain reliable uncertainty estimates under our formulation, we examine existing approaches, including Bayesian methods~\cite{torres2019surveybayesiannetworksapplications}, ensembles~\cite{10879299}, Kalman filtering~\cite{klein2022uncertaintydatadrivenkalmanfiltering}, and uncertainty embedding~\cite{oh2019modelinguncertaintyhedgedinstance}.
However, these methods often produce uncertainty estimates that do not consistently reflect predictive correctness, resulting in miscalibration~\cite{ghoshal2022calibratedmodeluncertaintydeep}.
Moreover, inaccurate uncertainty estimates exacerbate temporal boundary misalignment and lead to fragmented predictions, resulting in false alarms, missed risks, and unnecessary or delayed braking responses.
%
% Such instability can trigger unnecessary or delayed braking responses, ultimately degrading the safety of IDS~\cite{nhtsa2023humanfactorsdesignguidancefordrivervehicleinterfaces}.

%

%%% We propose Conformal Risk Tube Prediction Framework %%%
To this end, we present \textbf{Conformal Risk Tube Prediction}, a framework that integrates Conformal Prediction (CP)~\cite{shafer2007tutorialconformalprediction,angelopoulos2023conformalpidcontroltime} to construct calibrated risk tubes capturing both the spatial and temporal uncertainty of potential hazards.
%
% Conformal Prediction (CP)~\cite{shafer2007tutorialconformalprediction,angelopoulos2023conformalpidcontroltime} provides a principled framework for constructing calibrated predictive sets with guaranteed coverage of the true target. 
%
% Its informativeness is reflected in set size: smaller sets indicate higher confidence, while larger sets signal greater ambiguity~\cite{cresswell2024conformalpredictionsetsimprove}.
%
However, vanilla CP is insufficient in our setting because different risk categories (such as occlusion-induced or interaction-driven risks) exhibit distinct spatiotemporal characteristics that complicate calibration and reduce reliability.
To address this challenge, we introduce a spatiotemporal feature-alignment loss that encourages category-consistent appearance–motion representations. We further employ category-aware conformal calibrators to ensure reliable risk score calibration and predictive coverage across heterogeneous risk types.
% , enabling robust uncertainty-aware risk tube prediction in complex driving environments.
%
% To address this, we introduce a spatiotemporal feature-alignment loss to capture shared appearance–motion patterns within each category, and employ category-aware conformal calibrators to ensure reliable risk score calibration and coverage across heterogeneous risks.
% a novel conformal prediction framework that reduces confused calibration and localizes risks more precisely in space and time.

%%% Why propose new dataset? %%%
To evaluate our approach, we construct a \textbf{Multiple Coexisting Risks} dataset, in which multiple risk categories occur within a single scenario, a setting that is not addressed in the existing datasets~\cite{fang2023dadadriverattentionprediction,you2020CTA,kung2024riskbenchscenariobasedbenchmarkrisk,wang2023driveanywheregeneralizableendtoend,zhan2019interactiondatasetinternationaladversarial,shao2023reasonnetendtoenddrivingtemporal,nataly2024predictionoccludedpedestriansroad,xu2022safebenchbenchmarkingplatformsafety,wang2023deepaccidentmotionaccidentprediction,karim_am_net2023}. Our dataset enables comprehensive evaluation under multi-risk conditions.
We systematically analyze factors that influence uncertainty estimation, including scenario configurations, category-specific behaviors, and the propagation of perception errors, to assess the robustness of our method.
Experimental results demonstrate clear improvements over prior uncertainty-modeling baselines, achieving higher calibrated risk coverage, better temporal alignment, and fewer fragmented predictions.
Furthermore, we show that risk tubes enable timely yet minimal braking alerts~\cite{nhtsa2023humanfactorsdesignguidancefordrivervehicleinterfaces}, outperforming existing Vision-ROI methods.
Our contributions are summarized as follows:
%
% we review existing risk identification datasets~\cite{fang2023dadadriverattentionprediction,you2020CTA,kung2024riskbenchscenariobasedbenchmarkrisk,wang2023driveanywheregeneralizableendtoend,zhan2019interactiondatasetinternationaladversarial,shao2023reasonnetendtoenddrivingtemporal,nataly2024predictionoccludedpedestriansroad,xu2022safebenchbenchmarkingplatformsafety,wang2023deepaccidentmotionaccidentprediction,karim_am_net2023}.
% Although they cover diverse risk types, few consider concurrent risks (Table~\ref{tab:datasets-risk-types}), limiting analysis of compounded uncertainty.
%
%%% Demostrate the effectiveness of Risk Tube in Vision-ROI and Downstream Task %%%
%%% Summarize Contributions %%%
% \noindent 
\begin{itemize}
\item We introduce an uncertainty-aware Vision-ROI formulation, \textbf{Conformal Risk Tube Prediction} that models spatiotemporal uncertainty of potential hazards more reliably than existing approaches.

% \item We develop a \textbf{Conformal Risk Tube Prediction framework} that provides calibrated and consistent risk estimates under distribution shift and perception errors.

\item We construct a \textbf{Multiple Coexisting Risks dataset} that enables systematic evaluation of concurrent multi-risk scenarios, a setting that challenges existing Vision-ROI methods.

\item Extensive experiments demonstrate that our framework improves the robustness of Vision-ROI and supports more reliable downstream responses, such as reducing nuisance braking alerts.

\end{itemize}
\section{RELATED WORK}

\subsection{Visual Risk Object Identification}

Visual risk object identification (Vision-ROI) is a core capability of intelligent driving systems that aim to reduce accident frequency and severity.
Prior works can be categorized into four paradigms.
First, objects predicted to be involved in collisions are treated as risk objects~\cite{herzig2019spatiotemporalactiongraphnetworks,fang2023dadadriverattentionprediction,you2020CTA,Chandra_2019,malla2020titanfutureforecastusing,9577864}. 
Second, risk objects are defined by human annotators’ subjective assessments~\cite{10.1007/978-3-540-88682-2_40,OHNBAR2017425,zeng2017agentcentricriskassessmentaccident,gao2019goalorientedobjectimportanceestimation,li2022importantobjectidentificationsemisupervised}.
Third, objects fixated by human gaze are considered risk objects~\cite{7789504, xia2018predictingdriverattentioncritical,pal2020lookingrightstuffguided,baee2021medirlpredictingvisualattention}. 
Fourth, objects influencing the driver’s or the ego vehicle’s behavior are labeled as risk objects~\cite{li2020learning3dawareegocentricspatialtemporal,li2020makedriversstopdrivercentric,Gupta_2024,10177972,xiao2023learningroadscenelevelrepresentations,pao2025potentialfieldsceneaffordance}.

In this work, we focus on object importance–based Vision-ROI methods, which are typically deterministic and ignore predictive uncertainty. Such overconfident outputs~\cite{gawlikowski2022surveyuncertaintydeepneural} may compromise safety~\cite{Khaitan2020SafePA,9815528}.
Existing uncertainty-aware vision methods introduce error intervals~\cite{Bao_2020}, predefined candidate sets~\cite{huang2019uncertaintyawaredrivertrajectoryprediction}, or variance heat maps~\cite{9607788}, yet they lack a principled mechanism to model uncertainty that jointly evolves across space and time.
We therefore propose Risk Tube Prediction, an uncertainty-aware formulation that jointly models uncertainty over spatial extent and temporal horizon.
By marginalizing variability in both location and timing, it yields more robust risk estimates.
\vspace{-3pt}
\subsection{Uncertainty Quantification}
Uncertainty quantification enables driving systems to identify when predictions are unreliable~\cite{10879299}. 
In driving applications, methods generally fall into two families: direct modeling and statistical approaches. 
Direct modeling approaches~\cite{Bao_2020, Pitropov_2022, kendall2016bayesiansegnetmodeluncertainty, Tang_2019, Jospin_2022, klein2022uncertaintydatadrivenkalmanfiltering, 9607788} include Bayesian formulations~\cite{Jospin_2022} treat network weights as random variables and estimates predictive uncertainty via posterior sampling or variational approximations.
Kalman Filter-based method~\cite{klein2022uncertaintydatadrivenkalmanfiltering} that provides state uncertainty through the state covariance in a Gaussian dynamical model.
Ensemble methods~\cite{9607788} train multiple networks with different initializations and interpret the dispersion of their predictions as uncertainty.
Uncertainty Embedding~\cite{oh2019modelinguncertaintyhedgedinstance} methods capture uncertainty by allowing each input embedding to occupy a distributional region in the latent space rather than a fixed point.
However, these approaches often degrade under distribution shift~\cite{10879299,torres2019surveybayesiannetworksapplications}, suffer calibration errors~\cite{ghoshal2022calibratedmodeluncertaintydeep}, incur high computational cost~\cite{9607788}, and yield unreliable test-time behavior.

Conformal Prediction (CP)~\cite{shafer2007tutorialconformalprediction, angelopoulos2023conformalpidcontroltime} is a widely used statistical inference that constructs prediction sets with coverage guarantees for the true target, while the set size provides an informative measure that dynamically reflects model uncertainty.
CP has already been applied to object detection~\cite{timans2024adaptiveboundingboxuncertainties}, multi-object tracking~\cite{su2024collaborativemultiobjecttrackingconformal} and trajectory prediction~\cite{chen2025conformaltrajectorypredictionmultiview}, where coverage is especially valuable for safety-critical driving.
We present, to our knowledge, the first application of CP to Vision-ROI.
However, vanilla CP is insufficient: traffic scenes contain heterogeneous risk categories (e.g., occlusion, interaction) whose distinct characteristics confound calibration.
We propose a category-aware CP framework with a spatiotemporal feature-alignment loss, achieving improved calibration and more precise risk localization.
\subsection{Dataset for Risk Identification}

Existing risk identification datasets and benchmarks adopt different definitions of risk and are evaluated under specific risk categories.
For example, prior studies have examined risk scenarios including occlusion (hidden hazards)~\cite{shao2023reasonnetendtoenddrivingtemporal,nataly2024predictionoccludedpedestriansroad,wang2023deepaccidentmotionaccidentprediction}, collision (forced crashes)~\cite{fang2023dadadriverattentionprediction,you2020CTA,kung2024riskbenchscenariobasedbenchmarkrisk,karim_am_net2023}, interaction (dynamic social events)~\cite{kung2024riskbenchscenariobasedbenchmarkrisk,zhan2019interactiondatasetinternationaladversarial,xu2022safebenchbenchmarkingplatformsafety}, and obstacle (static blockages)~\cite{kung2024riskbenchscenariobasedbenchmarkrisk,wang2023driveanywheregeneralizableendtoend,xu2022safebenchbenchmarkingplatformsafety}.
However, these scenarios rarely consider concurrent occurrence, limiting the evaluation of multi-risk coupling across categories, which ultimately complicates uncertainty estimation and risk assessment.

To address this gap, we construct the \textbf{Multiple Coexisting Risks} (MCR) dataset, integrating all four risk categories within shared scenarios. 
Within a single scenario, multiple risk categories can occur concurrently or in sequence. 
Built in CARLA~\cite{dosovitskiy2017carlaopenurbandriving}, MCR supports scripted hazard behaviors and controllable traffic density, providing approximately 1000 scenarios for comprehensive multi-risk evaluation.
\begin{table}[t]
\centering
\caption{Comparison of risk identification datasets by risk category and single- vs. multi-risk settings. Risk categories are Interaction (I), Collision (C), Obstacle (Obs), Occlusion (Occ), and Normal Driving (N).}
\small
\setlength{\tabcolsep}{4pt}
\resizebox{\linewidth}{!}{%
\begin{tabular}{lccccc c}
\toprule
\multirow{2}{*}{\textbf{Dataset}}
& \multicolumn{5}{c}{\textbf{Risk Category}}
& \multirow{2}{*}{\makebox[2.5cm][c]{\textbf{Single/Multiple}}} \\
& I & C & Obs & Occ & N & \\
\midrule
MCR (Ours)         & $\checkmark$ &$\checkmark$ &$\checkmark$& $\checkmark$ &  & M \\
RiskBench~\cite{kung2024riskbenchscenariobasedbenchmarkrisk}    & $\checkmark$  & $\checkmark$ & $\checkmark$ & & $\checkmark$ & S \\
DADA~\cite{fang2023dadadriverattentionprediction}          &              & $\checkmark$ &            &              &              & S \\
CTA~\cite{you2020CTA}           & $\checkmark$ & $\checkmark$ & $\checkmark$ &              &              & S \\
Drive Anywhere~\cite{wang2023driveanywheregeneralizableendtoend}&              &            &            & $\checkmark$ &              & S \\
INTERACTION~\cite{zhan2019interactiondatasetinternationaladversarial}   & $\checkmark$ &            &            &              &              & S \\
DOS~\cite{shao2023reasonnetendtoenddrivingtemporal}           &              &            &            & $\checkmark$ &              & S \\
OccluRoads~\cite{nataly2024predictionoccludedpedestriansroad}    &              &            &            & $\checkmark$ &              & S \\
SafeBench~\cite{xu2022safebenchbenchmarkingplatformsafety}     &              & $\checkmark$ &            &              & $\checkmark$ & S \\
DeepAccident~\cite{wang2023deepaccidentmotionaccidentprediction}  &              & $\checkmark$ &            & $\checkmark$ &              & S \\
ROL~\cite{karim_am_net2023}           &              &            & $\checkmark$ &              & $\checkmark$ & S \\
\bottomrule
\end{tabular}}
\label{tab:datasets-risk-types}
\end{table}

\section{The \textbf{Multiple Coexisting Risks} Dataset}
\begin{figure}[t] 
  \centering
  \includegraphics[width=\columnwidth]{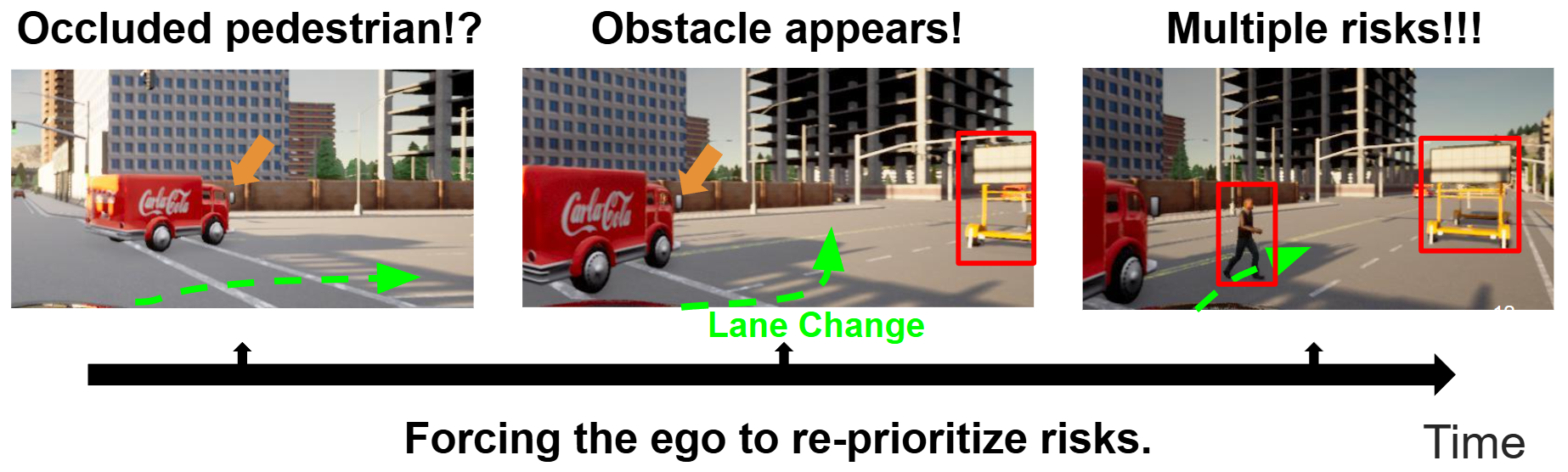}%
  \caption{The presence of multiple risks complicatedly reshapes object-ego interactions in both space and time.}
  \vspace{-13pt}
  \label{fig:multi_risk_scenario}
\end{figure}

\noindent We present a scenario taxonomy and data collection pipeline. Fig.~\ref{fig:multi_risk_scenario} shows an example of multi-risks scenarios.

\noindent\textbf{Scenario Taxonomy.}
\begin{figure}[t] 
  \centering
  \includegraphics[width=\columnwidth]{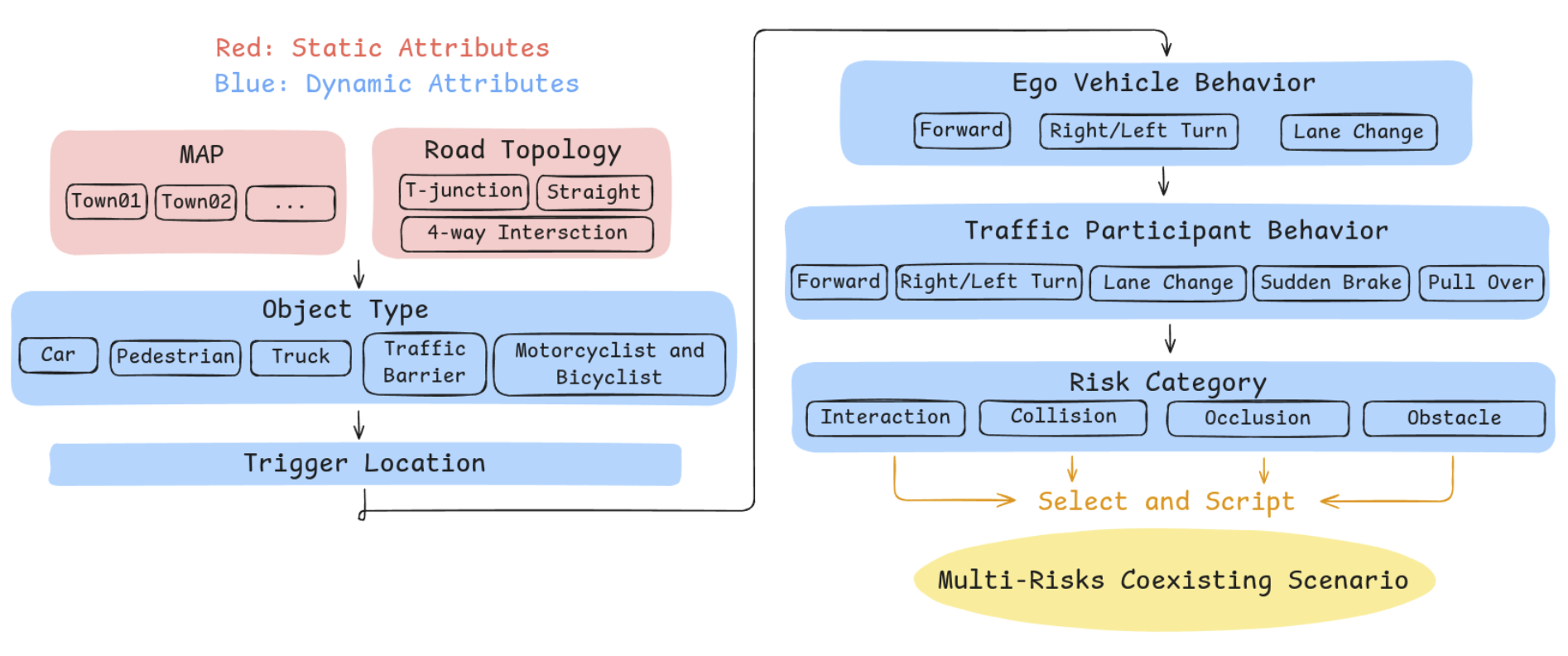} % 
  \caption{The scenario taxonomy specifies attributes including road topology, risk trigger location, risk category, object type, and behavior. Given a scenario configuration, we script hazard behaviors, control traffic conditions, and further augment the scenario by varying traffic density in CARLA.}
  \vspace{-12pt}
  \label{fig:scenario_taxonomy}
\end{figure}
We design a taxonomy (Fig.~\ref{fig:scenario_taxonomy}) with static (red) and dynamic (blue) attributes to systematically collect ground-truth risk instances from multiple coexisting categories.

\noindent \textbf{Static attributes} define the scene environment.
CARLA~\cite{dosovitskiy2017carlaopenurbandriving} provides towns with diverse layouts, e.g., Town02 (simple, many T-junctions) and Town05 (grid city with multi-lane intersections).
Following~\cite{kung2024riskbenchscenariobasedbenchmarkrisk}, the \textit{Map} selects the town, while \textit{Road Topology} specifies local structures (straight roads, T-junctions, four-way intersections).

\noindent \textbf{Dynamic attributes} determine object behavior patterns.
We first specify the \textit{Object Type} (e.g., motorcycle, car, pedestrian) to reflect agent heterogeneity.
The \textit{Risk Trigger Location} defines where the ego vehicle interacts with traffic participants, governing when and where risk materializes.
Varying this attribute generates diverse spatiotemporal configurations.
We design maneuver patterns for both the \textit{Ego Vehicle} and \textit{Traffic Participants}, including forward motion, lane changes, turns, and sudden braking.
We consider four \textit{Risk Categories} (Interaction, Collision, Obstacle, and Occlusion), each with distinct spatiotemporal characteristics.
A single scenario may include multiple categories by configuring each risk instance and composing them within the same scene.

\noindent\textbf{Data Collection.}
We use the Scenario Runner API in CARLA~\cite{dosovitskiy2017carlaopenurbandriving} to script scenarios, specify trigger locations, and instantiate object types.
Traffic participants follow interpolated trajectories between predefined start and end points.
The ego vehicle is controlled by a rule-based planner~\cite{shao2022safetyenhancedautonomousdrivingusing} that performs obstacle avoidance, lane changes, and speed control according to the script.
Additional random actors are spawned to increase environmental complexity and encourage interactions. 
Data are collected at 4 FPS, including RGB images and CARLA-provided metadata such as bounding boxes and velocities.
Since risk evolves over time with rising and falling phases, temporal annotations are manually provided.
In total, we generate approximately 1000 diverse scenarios with a balanced distribution of risk categories, enabling comprehensive evaluation under multi-risk settings.

\section{PRELIMINARIES}

\begin{figure*}[t!]
  \centering
  \includegraphics[width=1.0\textwidth]{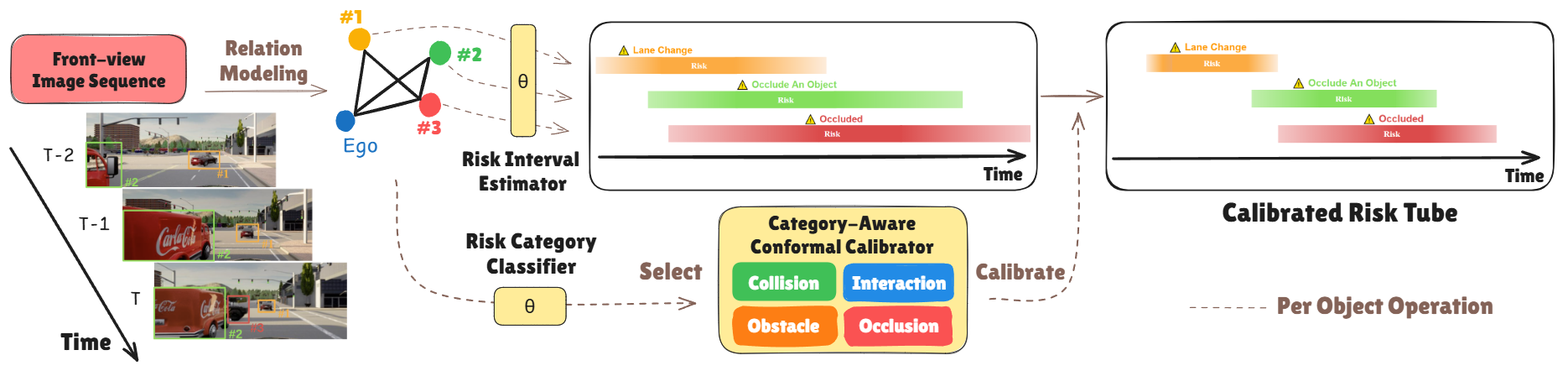}
  \caption{Overview of the Conformal Risk Tube Prediction Framework. Given front-view images, the model performs spatiotemporal relation modeling and predicts each object’s future risk interval. Then, based on a prediction of an object’s risk category, the corresponding conformal calibrator is applied to calibrate its risk scores over the future interval. The calibrated risk tube provides a more precise temporal bound to fully cover the true risk interval of each hazardous object.}
  \label{fig:conformal_risktube_framework}
  % \vspace{-10pt} 
\end{figure*}

\subsection{Problem Formulation of Risk Tube Prediction}

Given front-view image frames, the model outputs a \textbf{Risk Tube Prediction} $\mathcal{T}$ which encloses the set of uncertain risk objects along with the future time intervals during which each object may become hazardous.
\[
\mathcal{T}_{t}
\;=\;
\Bigl\{\,\bigl(o,\,[\hat t^{\mathrm{start}}_{o},\,\hat t^{\mathrm{end}}_{o}]\bigr)
\;:\;
o \in \hat{\mathcal{O}}_{t}\,\Bigr\}
\]
%
% This task is challenging, as the tube size must 
The predicted tube size should effectively reflect predictive uncertainty under diverse scenarios with multiple interacting risks.
Ideally, a smaller tube indicates higher model confidence, precisely localizing the risk within a tighter interval, whereas higher uncertainty results in a larger tube.
For evaluation, we consider an online setting, where the input consists of the past three front-view frames $I_{T-2:T}$.
The tube’s temporal support starts at $t=T$ and extends up to $t=T+7$ (an 8-step horizon, $H=8$).

\subsection{Conformal Prediction (CP)}
Consider a base predictor $f_\theta:\mathcal{X}\!\to\!\mathcal{Y}$ trained on a dataset $D_{\text{train}}$, for an input $x\in\mathcal{X}$, the model predicts $\hat{y}=f_\theta(x)\in\mathcal{Y}$ in the task-specific output.
CP constructs a \emph{prediction set} $C(X_{\text{test}})\subseteq\mathcal{Y}$ for a new sample $X_{\text{test}}$.
First, we define a nonconformity score $S(X,Y):\mathcal{X}\times\mathcal{Y}\to\mathbb{R}$ from the model’s outputs that quantifies how inconsistent the prediction is with the ground truth.
For classification, a common choice is
$S(x,y) \;=\; 1 - f_\theta(x)_y$,
where \(f_\theta(x)_y\) denotes the predicted probability of the true label \(y\).
For regression, we usually design $S(x,y) \;$as$\; \lvert y - f_\theta(x) \rvert.$
Given a calibration set $D_{\text{cal}}=\{(X_i,Y_i)\}_{i=1}^n$ of previously unseen, exchangeable pairs drawn from the same distribution as $D_{\text{train}}$.
Compute nonconformity scores $s_i$ for every $(X_i,Y_i)\in D_{\text{cal}}$, and sort them in ascending order.
Then we take $\hat{q}_{1-\alpha}$ be the empirical $1-\alpha$ quantile of $\{s_i\}_{i=1}^n,$ which serves as the threshold indicating how much error is still acceptable.
Note that $\alpha\in(0,1)$ is the user-specified miscoverage level.
For a new instance $X_{\text{test}}$ (with unknown $Y_{\text{test}}$ at inference time), we construct a CP set
$C\!\left(X_{\text{test}}\right) \;=\; \bigl\{\, y \in \mathcal{Y} \;:
\; S\!\left(X_{\text{test}},y\right) \le \hat{q}_{1-\alpha} \,\bigr\}$ for classification or a CP interval $ \bigl[f_\theta(x) - \hat{q},\; f_\theta(x) + \hat{q}\bigr]$ for regression.

Since the core assumption of CP is that samples are \emph{exchangeable}, the rank of the testing nonconformity score among the $n+1$ scores (n calibration plus itself) is uniformly distributed over $\{1,\dots,n\!+\!1\}$.  
If we choose \( k=\lceil (n+1)(1-\alpha)\rceil \) as the quantile \(\hat{q}\), then the probability that the test score falls within this quantile is $\frac{k}{n+1}\;$, i.e., coverage is guaranteed to be at least $1-\alpha$.
Therefore, the CP sets or intervals are designed to satisfy \emph{marginal coverage}, meaning they include the true label $Y_{\text{test}}$:
$\mathbb{P}\!\left\{\,Y_{\text{test}} \in C(X_{\text{test}})\,\right\}\;\ge\; 1-\alpha.$
We randomly split the dataset at the scenario level into disjoint training, calibration, and test sets following an 8:1:1 ratio, ensuring that no scenario appears in more than one split.

\section{METHODOLOGY}

\noindent 
The overall framework is shown in Fig.~\ref{fig:conformal_risktube_framework}.
Given an input image sequence, the base model extracts per-object features and produces future-interval risk scores forming a risk tube.
We train a risk-category classifier and apply a category-aware conformal predictor to calibrate the tube.
A spatiotemporal feature-alignment loss aligns features of objects within the same risk category across time and space.

\subsection{Base Model}\label{sec: base model}

Global ego features are extracted from RGB clips using an I3D backbone~\cite{carreira2018quovadisactionrecognition}.
RoIAlign~\cite{he2018maskrcnn} produces per-object features from detected boxes, with phantom boxes added for occluded regions to capture hidden objects.
A GCN~\cite{kipf2017semisupervisedclassificationgraphconvolutional} treats each object as a node and performs message passing with the ego node. 
Temporal relations are modeled via an LSTM-like module~\cite{6795963}.
A linear layer predicts each object’s risk score over eight timesteps, forming the deterministic risk tube, trained with binary cross-entropy on active-interval labels.
Conformal prediction is applied afterward for uncertainty estimation and calibration.

\subsection{Category-Aware Conformal Calibrator}\label{sec: CCAC}

Vanilla CP is insufficient as risk categories differ in spatiotemporal signatures, confusing calibration.
We train an MLP classifier to assign each object to a category (occlusion, obstacle, interaction, collision) and maintain a dedicated conformal calibrator per category~\cite{bhatnagar2023improvedonlineconformalprediction}.

Let $f_\theta(x)\in[0,1]^{H=8}$ denote the model's predicted 8 timesteps risk score interval.
For each timestep $t$, the model produces $\mathrm{pred}_t=f_\theta(x)_t\in[0,1]$
, and the ground truth label is $g_t\in\{0,1\}$ indicating risk (1) or no risk (0). 
We define the nonconformity score
$S_t \;=\; \bigl|\,g_t - \mathrm{pred}_t\,\bigr|.$
For each risk category $c$ and prediction horizon step $t\in\{1,\ldots,8\}$, we compute nonconformity scores $\{S_t^{(i)}\}_{i=1}^{n_c}$, where $n_c$ denotes the number of calibration samples belonging to category $c$ in the calibration set $D_{cal}$.
Then we take the empirical $(1-\alpha)$-quantile:
$
\hat q^{(c)}_{t,\,1-\alpha}
\;=\;
\mathrm{Quantile}_{1-\alpha}\!\bigl(\{S_t^{(i)}\}_{i=1}^{n_c}\bigr).
$
At inference time, we define the buffer zone at time $t$
$
\mathrm{BZ}_t \;=\; \bigl[\,\hat q^{(c)}_{t,\,1-\alpha},\; 1-\hat q^{(c)}_{t,\,1-\alpha}\,\bigr].
$
We calibrate the tube as follows: 
$
\text{risk if } \mathrm{pred}_t \ge 1-\hat q^{(c)}_{t,\,1-\alpha}, \:
\text{no risk if } \mathrm{pred}_t \le \hat q^{(c)}_{t,\,1-\alpha},$
and treat $\mathrm{pred}_t\in\bigl(\hat q^{(c)}_{t,\,1-\alpha},\,1-\hat q^{(c)}_{t,\,1-\alpha}\bigr)$
as ambiguous buffer zone to mitigates oscillation near the decision boundary.
We continuously update the quantile online~\cite{bhatnagar2023improvedonlineconformalprediction}, which can be interpreted as dynamically adapting the buffer zone used for determining whether an object is risky.

\subsection{Spatiotemporal Feature Alignment}

Objects in the same risk category often share appearance and motion patterns.
To encourage features of such objects to align in space in a manner that is
consistent with how their states evolve over time, we define a spatiotemporal alignment loss. 
Let $F_i(t)\in\mathbb{R}^H$ denote the latent node feature with length $H$ of object $i$ at time $t$.
For pairs $(i,k)$ that belong to the \emph{same} risk category (with $i\neq k$), we measure their \emph{spatial} similarity at time $t$ via cosine similarity, and we measure each object’s \emph{temporal} similarity between $t$ and $t{+}1$. 
We penalize the mismatch between the spatial similarity and the temporal similarity, averaged over a set $\mathcal{P}$ of valid triplets $(t,i,k)$.
\begin{itemize}
\medskip
\item Spatial Similarity:
\begin{equation}
\label{eq:spatial-cos}
\begin{aligned}
\cos_{\mathrm{spat}}^{(t,i,k)}
&=\frac{\langle F_i(t),\, F_k(t)\rangle}{\|F_i(t)\|\,\|F_k(t)\|},\\
&\text{for } k\neq i \quad \text{and }\mathrm{risk\_type}_k=\mathrm{risk\_type}_i 
\end{aligned}
\end{equation}

% --- Temporal Similarity ---
\item Temporal Similarity:
\begin{equation}
\label{eq:delta-temp}
\Delta_{\mathrm{temp}}^{(t,i,k)}
=
\frac{\langle F_i(t),\, F_i(t{+}1)\rangle}{\|F_i(t)\|\,\|F_i(t{+}1)\|}
-
\frac{\langle F_k(t),\, F_k(t{+}1)\rangle}{\|F_k(t)\|\,\|F_k(t{+}1)\|}
\end{equation}

% --- Alignment loss (unchanged) ---
\item Alignment Loss:
\begin{equation}
\label{eq:align-loss}
\mathcal{L}_{\mathrm{align}}
=
\frac{1}{|\mathcal{P}|}
\sum_{(t,i,k)\in\mathcal{P}}
\Bigl(
\cos_{\mathrm{spat}}^{(t,i,k)} - \Delta_{\mathrm{temp}}^{(t,i,k)}
\Bigr)^{2}
\end{equation}
\end{itemize}

\section{EXPERIMENTS}
\noindent Our experiments aim to answer the following research questions (RQ).
(\textbf{RQ1}) Is Conformal Risk Tube Prediction robust to spatiotemporal variations across risk categories?
(\textbf{RQ2}) Does it remain robust under propagated perception errors?
(\textbf{RQ3}) How does the Risk Tube benefit downstream tasks compared with other Vision-ROI methods?

\subsection{Baselines} \label{sec: baselines}

All baselines share the same base model (Sec.~\ref{sec: base model}) and use their estimated normalized uncertainty to construct buffer zones (Sec.~\ref{sec: CCAC}) for fair comparison on RQ1 and RQ2.
\textbf{Rule-based:} Every object is marked \emph{risky} at all time steps.
\textbf{HD (Hard Decision):} Deterministic classifier using a fixed threshold to label each risk interval element.
\textbf{BNN~\cite{Bao_2020}:} Bayesian neural network estimates predictive uncertainty via posterior sampling; variance across MC samples is used as uncertainty.
\textbf{KF~\cite{7528889}:} Kalman Filter provides state uncertainty through the state covariance; magnitude is used as uncertainty.
\textbf{OCP~\cite{bhatnagar2023improvedonlineconformalprediction}:} Online conformal prediction updates quantiles adaptively to provide coverage guarantees over time.
\textbf{UE~\cite{oh2019modelinguncertaintyhedgedinstance}:} Uncertainty Embedding maps features to a Gaussian latent $(\mu,\sigma^2)$ and trains with task loss plus KL penalty, capturing feature uncertainty.

For RQ3, we compare the following Vision-ROI methods using the same backbone.
\textbf{Distance:} Object is risky if distance to ego is below $10$m.
\textbf{Collision Anticipation (CA)~\cite{zeng2017agentcentricriskassessmentaccident}:} This method predicts which object will be involved in a collision; we use the predicted collision score as each object's risk score.
\textbf{Behavior Prediction (BP):}~\cite{li2020learning3dawareegocentricspatialtemporal}
The approach outputs the ego vehicle's current action (go/stop). When the predicted action is \emph{stop}, the object with the highest attention score is designated as the risk object.
\textbf{Trajectory Prediction (TP):}~\cite{Chandra_2019}
We predict future 2D trajectories on image for objects and mark an object as risky if its predicted trajectory intersects the ego vehicle's path.

\subsection{Evaluation Metrics}
We first describe metrics for RQ1 and RQ2.

\noindent \textbf{Coverage:}
The ratio of GT risk objects whose \emph{active risk interval} is fully covered by the prediction (equivalently, the GT interval is a subset of the predicted interval).

\noindent \textbf{Tube Volume (TV):}
Serving as an indicator of predictive uncertainty.
Formally, for object $o$ with predicted risk interval $\widehat{\mathcal{I}}_o$ in the tube,
$
\mathrm{TV} \;=\; \frac{1}{|\widehat{\mathcal{O}}|}\sum_{o\in \widehat{\mathcal{O}}} \lvert \widehat{\mathcal{I}}_o \rvert.
$
At a fixed coverage level, larger TV implies greater uncertainty.

\noindent \textbf{Temporal Consistency (TC):}
Quantifies fragmented prediction.
We define the number of temporal switches
$
\mathrm{T}(y) \;=\; \sum_{t=0}^{H-1} \mathbf{1}\!\left[y_t \ne y_{t+1}\right],
(H=8).
$
Temporal consistency compares the switch counts of prediction and ground truth:
$
\mathrm{TC}
\;=\;
1 \;-\; \frac{\bigl|\,\mathrm{T}(\widehat{\mathcal{I}}_o^{\text{pred}})-\mathrm{T}(\widehat{\mathcal{I}}_o^{\text{gt}})\,\bigr|}{H-1}.
$
Higher values indicating closer temporal behavior to ground truth.

\noindent \textbf{Boundary Alignment (BA):}
Quantifies temporal boundary misalignment.
We evaluate alignment near the
\emph{risk start} boundary $T_s$  and the \emph{risk end} boundary $T_e$.
Let $m(t) = \mathbf{1}\{ \text{pred}(t) = \text{gt}(t) \} \in \{0,1\}$ be the per–time-step match indicator.
Let $w_\theta(t)=\exp(-|t-\theta|/\tau)$ be the penalty weights around a boundary $\theta\in\{T_s,T_e\}$, where $\tau$ controls how fast the penalty decays as $t$ moves away from the boundary.
The boundary score is the locally weighted accuracy around $\theta$:
$
\mathrm{PIC}^{*}_{\theta}
\;=\;
1 \;-\; \frac{\sum_t w_\theta(t)\,[1-m(t)]}{\sum_t w_\theta(t)}.
$
The final metric averages both sides:
$
\mathrm{BA}=\tfrac{1}{2}\bigl(\mathrm{PIC}^{*}_{T_s}+\mathrm{PIC}^{*}_{T_e}\bigr),
$
where larger is better.

\noindent \textbf{Risk-IOU:}
Let $\widehat{\mathcal{I}}$ and $\mathcal{I}^{\ast}$ be the predicted and
ground-truth active risk intervals, respectively. Define the interval IoU as
$
\mathrm{IoU}(\widehat{\mathcal{I}},\mathcal{I}^{\ast})
= \frac{\lvert\,\widehat{\mathcal{I}}\cap \mathcal{I}^{\ast}\,\rvert}
       {\lvert\,\widehat{\mathcal{I}}\cup \mathcal{I}^{\ast}\,\rvert}.
$
We combine it with Temporal Consistency (TC) and Boundary Alignment (BA) to obtain:
$
\mathrm{Risk IOU}
\;=\;
\mathrm{IoU}(\widehat{\mathcal{I}},\mathcal{I}^{\ast})
\;\times\;
\frac{\mathrm{TC} + \mathrm{BA}}{2}\,.
$

We then describe metrics for RQ3.

\noindent \textbf{Average Brake Counts:}
Measures how often the model triggers braking alerts on average.

\noindent \textbf{Misaligned Brake Counts (MBC):}
Quantify whether the braking timing is correct.
For a video clip (length=L) with predicted brake sequence $y_t\in\{0,1\}$ and ground-truth sequence $\hat{y}_t\in\{0,1\}$ over $t=1,\ldots,L$, the misaligned brake count is the sum of \emph{false negative brakes}  and
\emph{false positive brakes}:
$
\mathrm{MBC}
\;=\;
\sum_{t=1}^{L} \Big[ \mathbf{1}(y_t=0,\ \hat{y}_t=1) \;+\; \mathbf{1}(y_t=1,\ \hat{y}_t=0) \Big].
$

\subsection{Results and Discussions}

\begin{figure*}[!t]
  \centering
  \includegraphics[width=1.0\textwidth]{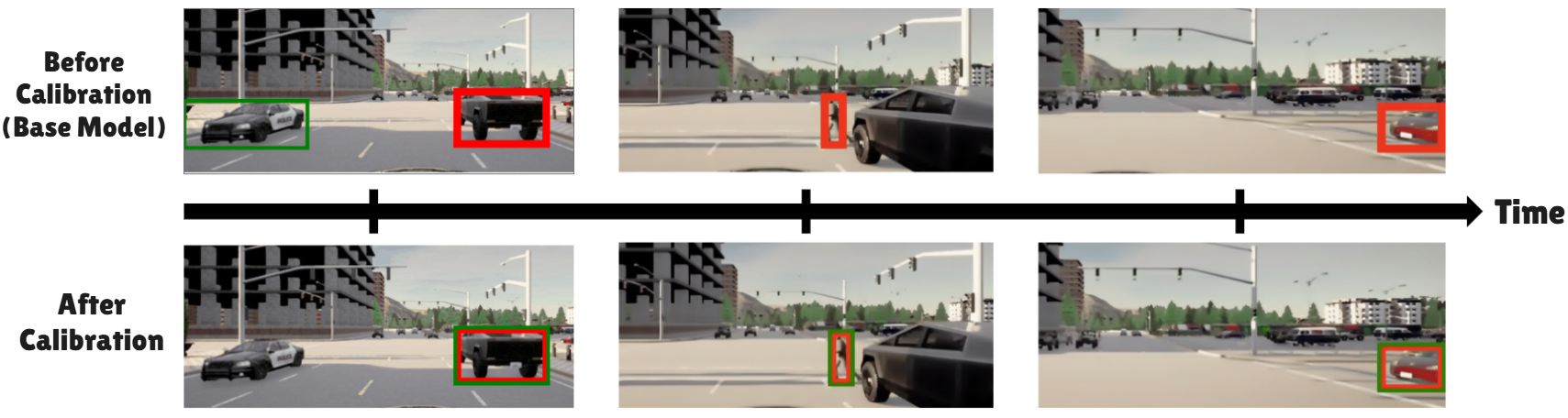}
  \caption{Visualization of ROI results before and after calibration on a sampled scenario. All detected risk objects are shown with green bounding boxes, while ground truth risks are in red.}
  \label{fig:vis_cal}
  \vspace{-10pt} 
\end{figure*}

\noindent \textbf{RQ1: Is Conformal Risk Tube Prediction robust to spatiotemporal variations across risk categories?}
We compare methods under both One-Risk and Multi-Risk settings in Table~\ref{tab:multi_risks_singlecol}.
The rule-based baseline trivially achieves Coverage = 1.0 by marking all timesteps as risky, inflating Tube Volume and yielding poor boundary alignment.
HD produces the smallest tubes but fails to cover full risk intervals, resulting in low coverage and overconfident predictions.
Other direct uncertainty modeling methods (BNN, KF, OCP, UE) also struggle.
In contrast, our method maintains high coverage with moderate Tube Volume, achieving strong boundary alignment and the best Risk-IoU in both settings.
Performance drops for all methods in the Multi-Risk setting, reflecting the increased difficulty under interacting risks.
Nevertheless, our approach remains robust due to improved uncertainty modeling and category-aware calibration, enabling more precise spatiotemporal localization (Fig.~\ref{fig:vis_cal}).

\begin{comment}
\begin{figure}[t] 
  \centering
  \includegraphics[width=\columnwidth]{figure/vis_more.png}%
  \caption{More visualizations on sampled scenarios.}
  \label{fig:vis_more}
  \vspace{-10pt}
\end{figure}
\end{comment}

We further analyze performance by risk category in Table~\ref{tab:per risk category analysis}.
While baselines improve only in specific categories, our method achieves consistently strong results across most metrics and categories, demonstrating robustness.
This result indicates that category-aware conformal calibration is beneficial.
All methods, including ours, perform worse in occlusion scenarios, likely because 2D phantom boxes overlap with foreground objects degrade feature quality.
This suggests the need for more dedicated occlusion modeling.

\begin{table}[t]
\centering
\caption{Multi-Risks Coupling Effects: results on \textbf{One Risk} and \textbf{Multi-Risks}. Higher is better for Coverage, Temporal Consistency (TC), Boundary Alignment (BA), and Risk IoU; lower is better for Tube Volume (TV). The best results are
highlighted in bold, and the second are underlined.}
\label{tab:multi_risks_singlecol}
\setlength{\tabcolsep}{4pt}
\footnotesize
\resizebox{\columnwidth}{!}{%
\begin{tabular}{llcccc|c}
\toprule
\textbf{Scenario} & \textbf{Method} & \textbf{Coverage} $\uparrow$ & \textbf{Tube Volume} $\downarrow$ & \textbf{TC} $\uparrow$ & \textbf{BA} $\uparrow$ & \textbf{Risk IoU} $\uparrow$ \\
\midrule
\multirow{7}{*}{\textbf{One Risk}}
  & Rule Based      & \textbf{1.000} & 23.261 & \textbf{0.857} & 0.600 & 0.475 \\
  & HD              & 0.667 & \textbf{7.432}   & 0.644 & 0.694 & 0.515 \\
  & BNN~\cite{Bao_2020} & 0.810 & 18.049  & 0.727 & 0.650 & 0.518 \\
  & KF~\cite{7528889}   & 0.778 & 14.560  & 0.716 & 0.720 & 0.529 \\
  & OCP~\cite{bhatnagar2023improvedonlineconformalprediction}
                     & 0.801 & 15.253  & 0.703 & \underline{0.751} & \underline{0.549} \\
  & UE~\cite{oh2019modelinguncertaintyhedgedinstance}  & 0.821 & 18.166 & 0.718 & 0.709 & 0.542 \\
  & \textbf{Ours}             & \underline{0.851} & \underline{12.988}  & \underline{0.734} & \textbf{0.800} & \textbf{0.637} \\
\midrule
\multirow{7}{*}{\textbf{Multi-Risks}}
  & Rule Based      & \textbf{1.000} & 28.844 & \textbf{0.857} & 0.582 & 0.494 \\
  & HD              & 0.625 & \textbf{6.507}    & 0.642 & 0.661 & 0.505 \\
  & BNN~\cite{Bao_2020} & 0.787 & 23.269  & 0.693 & 0.602 & 0.508 \\
  & KF~\cite{7528889}   & 0.715 & 18.005  & 0.704 & 0.654 & 0.519 \\
  & OCP~\cite{bhatnagar2023improvedonlineconformalprediction}
                     & 0.742 & 17.265   & 0.688 & \underline{0.665} & \underline{0.532} \\
  & UE~\cite{oh2019modelinguncertaintyhedgedinstance} & 0.798	& 18.300 & 0.707 & 0.664 & 0.527 \\
  & \textbf{Ours}             & \underline{0.827} & \underline{15.641}   & \underline{0.708}& \textbf{0.752} & \textbf{0.569} \\
\bottomrule
\vspace{-20pt}
\end{tabular}%
}
\end{table}

\begin{table}[t]
\centering
\caption{Per–Risk-Category Analysis.}
\label{tab:per risk category analysis}
\setlength{\tabcolsep}{4pt}   % 視需要再縮
\footnotesize                 % 或 \scriptsize 讓字更小
\resizebox{1.0\columnwidth}{!}{%
\begin{tabular}{llcccc|c}
\toprule
\textbf{Category} & \textbf{Method} & \textbf{Coverage} $\uparrow$ & \textbf{TV} $\downarrow$ & \textbf{TC} $\uparrow$ & \textbf{BA} $\uparrow$ & \textbf{Risk IoU} $\uparrow$ \\
\midrule
\multirow{6}{*}{\textbf{Interaction}}
 & HD   & 0.615 & 8.101  & 0.652 & 0.756 & 0.578 \\
 & BNN~\cite{Bao_2020}  & 0.792 & 21.652 & 0.757 & 0.817 & 0.625 \\
 & KF~\cite{7528889}   & 0.816 & 18.014 & 0.695 & 0.837 & 0.623 \\
 & OCP~\cite{bhatnagar2023improvedonlineconformalprediction}  & 0.842 & 17.796 & 0.723 & \textbf{0.839} & 0.655 \\
 & UE~\cite{oh2019modelinguncertaintyhedgedinstance} & 0.823 & 18.618 &0.766 &0.811	& 0.643 \\
 & \textbf{Ours} & \textbf{0.876} & 14.458 & \textbf{0.796} & 0.826 & \textbf{0.681} \\
\midrule
\multirow{6}{*}{\textbf{Collision}}
 & HD   & 0.609 & 6.646  & 0.656 & 0.698 & 0.516 \\
 & BNN~\cite{Bao_2020}  & 0.833 & 22.039 & 0.657 & 0.734 & 0.534 \\
 & KF~\cite{7528889}   & 0.825 & 20.014 & 0.695 & 0.792 & 0.585 \\
 & OCP~\cite{bhatnagar2023improvedonlineconformalprediction}  & 0.847 & 18.699 & 0.732 & 0.816 & 0.610 \\
 & UE~\cite{oh2019modelinguncertaintyhedgedinstance} & 0.833 &19.489 &0.712 &0.798 & 0.602 \\
 & \textbf{Ours} & \textbf{0.865} & 14.839 & \textbf{0.788} & \textbf{0.839} & \textbf{0.674} \\
\midrule
\multirow{6}{*}{\textbf{Occlusion}}
 & HD   & 0.661 & 7.996  & 0.654 & 0.648 & 0.508 \\
 & BNN~\cite{Bao_2020}  & 0.811 & 24.681 & 0.757 & 0.721 & 0.532 \\
 & KF~\cite{7528889}   & 0.813 & 24.016 & 0.695 & 0.747 & 0.513 \\
 & OCP~\cite{bhatnagar2023improvedonlineconformalprediction}  & 0.791 & 18.025 & \textbf{0.786} & 0.788 & 0.589 \\
 & UE~\cite{oh2019modelinguncertaintyhedgedinstance} & 0.806 & 22.486 &0.753 &0.756 & 0.571 \\
 & \textbf{Ours} & \textbf{0.828} & 16.807 & 0.766 & \textbf{0.807} & \textbf{0.604} \\
\midrule
\multirow{6}{*}{\textbf{Obstacle}}
 & HD   & 0.728 & 10.359 & 0.751 & 0.680 & 0.570 \\
 & BNN~\cite{Bao_2020}  & 0.802 & 19.879 & 0.757 & 0.714 & 0.581 \\
 & KF~\cite{7528889}   & 0.785 & 18.450 & 0.719 & 0.796 & 0.628 \\
 & OCP~\cite{bhatnagar2023improvedonlineconformalprediction}  & 0.821 & 18.014 & 0.695 & 0.764 & 0.595 \\
 & UE~\cite{oh2019modelinguncertaintyhedgedinstance} & 0.811 & 17.540 &0.724 & 0.783	& 0.638 \\
 & \textbf{Ours} & \textbf{0.847} & 15.133 & \textbf{0.779} & \textbf{0.828} & \textbf{0.682} \\
\bottomrule
\end{tabular}%
}
\vspace{-5pt}
\end{table}

\noindent \textbf{RQ2: Does the Conformal Risk Tube Prediction remain robust under propagated perception errors?}
Table~\ref{tab:perception error} shows that replacing ground-truth boxes with perception-based detections degrades all methods: Tube Volume increases, while Coverage, TC, BA, and Risk-IoU decrease.
Despite this, our method remains the most robust, exhibiting the smallest Tube Volume inflation and the lowest performance drops across other metrics, whereas baselines suffer substantially larger penalties. 
Our approach enlarges tubes only as needed to absorb detection noise while better preserving temporal fidelity.
Notably, our current formulation models spatial uncertainty at the object level (identity) rather than explicitly at the bounding-box level.
The performance gap under detection inputs reveals the impact of spatial uncertainty in real perception pipelines, suggesting that finer region-level uncertainty modeling could further enhance robustness.
% —an important direction for future work.
\\

\begin{table}[t]
\centering
\caption{Each entry shows the change in metrics when replacing \textbf{GT} bounding boxes with \textbf{perception} bounding boxes ($\Delta=\text{Detected bbox result}-\text{GT bbox result}$).}
\label{tab:perception error}
\setlength{\tabcolsep}{4pt}
\footnotesize
\resizebox{\columnwidth}{!}{%
\begin{tabular}{llcccc|c}
\toprule
\textbf{Scenario} & \textbf{Method} &
$\Delta$\textbf{Coverage} $\uparrow$ &
$\Delta$\textbf{TV} $\downarrow$ &
$\Delta$\textbf{TC} $\uparrow$ &
$\Delta$\textbf{BA} $\uparrow$ &
$\Delta$\textbf{Risk IoU} $\uparrow$ \\
\midrule
\multirow{6}{*}{\textbf{One Risk}}
  & HD   & -0.102 & +4.921 & -0.143 & -0.137 & -0.150 \\
  & BNN~\cite{Bao_2020}  & -0.078 & +3.361 & -0.170 & -0.118 & -0.142 \\
  & KF~\cite{7528889}   & -0.159 & +4.412 & -0.180 & -0.141 & -0.149 \\
  & OCP~\cite{bhatnagar2023improvedonlineconformalprediction}  & -0.088 & +3.759 & -0.136 & -0.162 & -0.148 \\
  & UE~\cite{oh2019modelinguncertaintyhedgedinstance} & -0.083 & +3.518 &-0.114&	-0.128	&-0.139 \\
  & \textbf{Ours} & \textbf{-0.071} & \textbf{+3.021} & \textbf{-0.094} & \textbf{-0.100} & \textbf{-0.130} \\
\midrule
\multirow{6}{*}{\textbf{Multi-Risks}}
  & HD   & -0.171 & +5.940 & -0.224 & -0.182 & -0.208 \\
  & BNN~\cite{Bao_2020}  & -0.134 & +6.207 & -0.197 & -0.175 & -0.189 \\
  & KF~\cite{7528889}   & -0.237 & +5.956 & -0.201 & -0.172 & -0.183 \\
  & OCP~\cite{bhatnagar2023improvedonlineconformalprediction}  & -0.138 & +5.655 & -0.185 & -0.161 & -0.178 \\
  & UE~\cite{oh2019modelinguncertaintyhedgedinstance} & -0.144 & +5.567 & -0.181 &-0.164 &-0.172 \\
  & \textbf{Ours} & \textbf{-0.104} & \textbf{+5.180} & \textbf{-0.158} & \textbf{-0.136} & \textbf{-0.148} \\
\bottomrule
\end{tabular}%
}
% \vspace{-10pt}
\end{table}

\noindent \textbf{RQ3: How does the Risk Tube benefit downstream tasks compared with other Vision-ROI methods?}
An intelligent driving system must react promptly to hazards to prevent accidents.
We use braking alerts as the system’s response mechanism.
Triggering brakes solely based on object–ego distance produces frequent nuisance alerts.
Incorporating Vision-ROI allows the system to focus on truly hazardous objects and suppress spurious triggers.
We evaluate multiple Vision-ROI methods (Sec.~\ref{sec: baselines}) and our Risk Tube Prediction on downstream braking using Average Brake Count and Misaligned Brake Count.
As shown in Table~\ref{tab:brake_onecol_scenario_left}, combining distance proximity (e.g., $<10$,m) with Vision-ROI risk substantially reduces nuisance alerts.
Notably, the Risk Tube achieves the lowest brake counts and misalignment.
By providing calibrated estimates of when risk begins and ends, it minimizes unnecessary interventions while closely matching ground-truth braking behavior.
Overall, calibrated risk tubes serve as a principled gating mechanism, reducing nuisance braking without compromising safety.

% preamble: \usepackage{booktabs,multirow,graphicx}
\begin{table}[t]
\centering
\caption{Downstream braking alerts under the criteria
“Vision-ROI flags risky \emph{and} distance $<10$\,m.”
Lower is better for both metrics. GT gives the empirical lower bound on brake counts; MBC is undefined for GT (shown as “—”).}
\label{tab:brake_onecol_scenario_left}
\setlength{\tabcolsep}{4pt}
\footnotesize
\resizebox{\columnwidth}{!}{%
\begin{tabular}{llcc}
\toprule
\textbf{Scenario} & \textbf{Method} & \textbf{Average Brake Counts} $\downarrow$ & \textbf{Misaligned Brake Counts} $\downarrow$ \\
\midrule
\multirow{6}{*}{\textbf{One Risk}}
  & Ground Truth            & 16.22 & ---   \\
  & Distance ($<10$ m)      & 43.26 & 29.47 \\
  & CA~\cite{zeng2017agentcentricriskassessmentaccident}  & 32.17 & 21.43 \\
  & BP~\cite{li2020learning3dawareegocentricspatialtemporal}     & 31.09 & 20.09 \\
  & TP~\cite{Chandra_2019}   & 35.48 & 23.43 \\
  & \textbf{Risk Tube (Ours)} & \textbf{23.65} & \textbf{16.34} \\
\midrule
\multirow{6}{*}{\textbf{Multi-Risks}}
  & Ground Truth            & 21.61 & ---   \\
  & Distance ($<10$ m)      & 54.28 & 37.40 \\
  & CA~\cite{zeng2017agentcentricriskassessmentaccident}  & 40.78 & 28.98 \\
  & BP~\cite{li2020learning3dawareegocentricspatialtemporal}     & 36.87 & 27.08 \\
  & TP~\cite{Chandra_2019}   & 36.44 & 26.05 \\
  & \textbf{Risk Tube (Ours)} & \textbf{28.41} & \textbf{20.68} \\
\bottomrule
\end{tabular}%
}
\end{table}

\subsection{Ablation Study}

\noindent We justify the design choices built upon the base model, with results presented in Table~\ref{tab:ablation}.
Adding STFA improves Coverage, TC, and BA, while reducing TV compared with the Base model, indicating that aligning object features across space and time yields more stable risk intervals and fewer unnecessary expansions.
Introducing CACC on top of STFA yields further improvements, achieving higher coverage, lower TV, and better Risk IoU compared to STFA alone.
These results demonstrate that category-aware calibration adjusts risk scores and uncertainty according to the characteristics of each risk category and helps maintain nominal coverage while producing more temporally aligned risk tubes.

\begin{table}[t]
\centering
\caption{Ablation study of model components. STFA denotes spatiotemporal feature alignment. CACC denotes category-aware conformal calibrators.}
\label{tab:ablation}
\setlength{\tabcolsep}{4pt}
\footnotesize
\resizebox{\columnwidth}{!}{%
\begin{tabular}{lccccc}
\toprule
\textbf{Method} & \textbf{Coverage} $\uparrow$ & \textbf{TV} $\downarrow$ & \textbf{TC} $\uparrow$ & \textbf{BA} $\uparrow$ & \textbf{Risk IoU} $\uparrow$ \\
\midrule
Base                     & 0.771 & 23.269 & 0.633 & 0.672 & 0.527 \\
Base + STFA               & 0.805 (+0.034) & 20.053 (-3.216) & 0.665 (+0.032) & 0.708 (+0.036) & 0.559 (+0.032) \\
Base + STFA + CACC (Ours) & \textbf{0.857 (+0.052)} & \textbf{15.641 (-4.412)} & \textbf{0.708 (+0.043)} & \textbf{0.732 (+0.024)} & \textbf{0.609 (+0.050)} \\
\bottomrule
\vspace{-18pt}
\end{tabular}%
}
\end{table}

\section{CONCLUSION}

We present Conformal Risk Tube Prediction, an uncertainty-aware formulation for Visual–ROI. 
We demonstrate that integrating conformal prediction can address temporal boundary misalignment, fragmented predictions, and miscalibrated uncertainty present in the existing object importance-based Vision-ROI algorithms.
Through our extensive experiments on the proposed Multiple Coexisting Risks dataset, we show that the proposed method is effective and robust across diverse scenario configurations.
% , risk categories with distinct characteristic and perception error.
%
Moreover, our method provides immediate yet minimal false alarms for downstream tasks such as braking warning.

\noindent \textbf{Limitations and Future Work.}
Performance under \textit{Occlusion} category remains weaker than the other risk types (Table~\ref{tab:per risk category analysis}).
We plan to conduct experiments on real-world settings.
Currently, risk tube predictions are made independently at each timestep; conditioning future predictions on past tubes could further improve temporal consistency.
Finally, we aim to extend our approach to other safety-critical tasks, such as lane-change avoidance and junction yielding, to enhance the generalizability of the proposed framework.

\noindent \textbf{Acknowledgment: }
The work is sponsored in part by the National Science and Technology Council under grants 113-2628-E-A49-022-, 114-2628-E-A49-007-, 114-2634-F-A49-004-, and the Ministry of Education, the Yushan Fellow Program Administrative Support Grant.

\bibliographystyle{IEEEtran}
\bibliography{root.bib}  % .bib

\end{document}